\documentclass[letterpaper, 10 pt, conference]{ieeeconf}
\IEEEoverridecommandlockouts 
\usepackage{nick_macros}
\usepackage{mathnotation}
\usepackage{float}
\usepackage{graphicx}
\usepackage{xcolor}
\usepackage{newtxtext}
\usepackage{booktabs} 
\usepackage{siunitx}  
\usepackage{soul}
\usepackage{cite}

\usepackage{etoolbox}
\usepackage{verbatim}
\newtoggle{showtodo}
\togglefalse{showtodo}  

\newenvironment{todolist}
  {\iftoggle{showtodo}{\begin{itemize}\color{red}}{\comment}}
  {\iftoggle{showtodo}{\end{itemize}}{\endcomment}}

\newcommand{\thrustang}{\beta}

\usepackage{amsthm}

\makeatletter
\let\NAT@parse\undefined
\makeatother
\usepackage[hidelinks]{hyperref}  

\title{\LARGE \bf Pose Estimation of a Thruster-Driven Bioinspired Multi-Link Robot*}

\author{{Nicholas B. Andrews}$^{1}$, {Yanhao Yang}$^{2}$, {Sofya Akhetova}$^{1}$, {Kristi A. Morgansen}$^{1}$, and {Ross L. Hatton}$^{2}$
\thanks{*This work was supported in part by {ONR Award N00014-23-1-2171}}%
\thanks{$^{1}$Department of Aeronautics and Astronautics, University of Washington, Seattle, WA, USA  
{\tt\small [{\href{mailto:nian6018@uw.edu}{nian6018}, \href{mailto:akhetova@uw.edu}{akhetova}, \href{mailto:morgansn@uw.edu}{morgansn}}]@uw.edu}}
\thanks{$^{2}$Collaborative Robotics and Intelligent Systems (CoRIS) Institute, Oregon State University, Corvallis, OR, USA {\tt\small [{\href{mailto:yangyanh@oregonstate.edu}{yangyanh}, \href{mailto:Ross.Hatton@oregonstate.edu}{ross.hatton}}]@oregonstate.edu}}
}

\begin{document}

\maketitle
\thispagestyle{empty}
\pagestyle{empty}

\begin{todolist}
    \item Please perform a close spelling review. The paper contains several sentences that require improvement and typos that need correction.
    \item \st{Typically, table captions are placed above the table.}
    \item \st{I recommend capitalizing the term "Unscented Kalman Filter (UKF)" throughout the manuscript, as it refers to a specific algorithm rather than a descriptive phrase. Using lowercase ("unscented") may cause confusion and could be misinterpreted outside the technical context.}
    \item The overall structure of the paper could be improved to enhance readability and logical flow between sections. For instance, Figure 8 is referenced in the Experimental Results Section, but appears in the middle of the Conclusions, which may confuse readers.
    \item remove equation numbers from equations not referenced
\end{todolist}

\begin{abstract}
This work demonstrates simultaneous pose (position and orientation) and shape estimation for a free-floating, bioinspired multi-link robot with unactuated joints, link-mounted thrusters for control, and a single gyroscope per link, resulting in an underactuated, minimally sensed platform. Because the inter-link joint angles are constrained, translation and rotation of the multi-link system requires cyclic, reciprocating actuation of the thrusters, referred to as a gait. Through a proof-of-concept hardware experiment and offline analysis, we show that the robot's shape can be reliably estimated using an Unscented Kalman Filter augmented with Gaussian process residual models to compensate for non-zero-mean, non-Gaussian noise, while the pose exhibits drift expected from gyroscope integration in the absence of absolute position measurements. Experimental results demonstrate that a Gaussian process model trained on a multi-gait dataset (forward, backward, left, right, and turning) performs comparably to one trained exclusively on forward-gait data, revealing an overlap in the gait input space, which can be exploited to reduce per-gait training data requirements while enhancing the filter's generalizability across multiple gaits. Lastly, we introduce a heuristic derived from the observability Gramian to correlate joint angle estimate quality with gait periodicity and thruster inputs, highlighting how control affects estimation quality.
\end{abstract}

\section{INTRODUCTION}
\begin{todolist}
    \item The problem under consideration should be presented in a broader view (to ensure a wider audience accessibility): the FloatSalp system should be introduced from the very beginning of the paper, explaining in which sense the experimental ground/wheeled setup corresponds to the real floating system; a comment on possible practical applications of this type of a robot is expected as well (i.e., better justification of the work is expected here).
    \item novelty and contribution of the paper could be better (more explicitly) highlighted
    \item The description of the innovations of this paper in the Introduction section is rather general. The authors should provide a detailed summary of the contributions of this paper at technical and theoretical levels to facilitate the readers' understanding.
    \item I think it would be worthwhile to briefly point out the practical applications of the class of robots considered in this paper.
    \item At the beginning of the Introduction, the paper states that the performance of dynamical systems critically depends on accurate knowledge of the system state to ensure robustness against disturbances and maintain safety guarantees. However, knowing the system state alone does not automatically ensure safety. The controller must also predict how the state will evolve under given inputs, decide on actions that keep the system within safe regions (through control policies, constraints, or safety filters), and react to uncertainties or estimation errors.
    \item In addition to the references, I recommend including a short description in the Introduction explaining what the FloatSalp is and what it is used for, rather than mentioning it only in the Conclusions.
\end{todolist}

In this paper, we investigate the specific challenge of pose and shape estimation for free-floating multi-link chains of rigid bodies in which joints are unactuated, each link is controlled through a thrust applied at its center of mass, and measurements come from gyroscopes mounted on each link, presenting an underactuated and minimally sensed multi-link system. These system constraints are motivated by our work on next-generation underwater autonomous vehicles inspired by colonial marine organisms called salps--jellyfish-like organisms that generate thrust in a single direction and form large chains with one another. Our bio-inspired robot design is composed of self-contained units that can mechanically connect and disconnect to form configurable chains, where the modular, minimally capable, and inexpensive nature of each unit allows the overall chain to be tailored for specific inspection or manipulation tasks. While this work is motivated by underwater robotics, the multi-link system considered here is broadly applicable to terrestrial or space-based chains of connected rigid bodies, such as tethered spacecraft.



Prior state estimation work for multi-link and snake robots has used Kalman filters to fuse information from multiple sensors into state and pose estimates~\cite{Rollinson2011-ti, Rollinson2016-kd, Rollinson2013-zx, Tully2011-ka}. However, the systems considered in these works differ from ours in two key aspects: they were equipped with joint angle encoders, and they were directly controlled via torque or position control at the joints, whereas our system has passive joints and is driven by thrusters equipped to each link. Other work has aimed at estimating the shape of multi-link systems using inertial measurement units (IMUs) without joint encoders~\cite{Furukawa2021-vb, Zhang2024-ws, Cox2023-mt}. However, these works have focused on uncontrolled, fixed-base chains over short time horizons, rather than the free-floating chains considered here, and therefore do not address simultaneous pose and shape estimation.

\begin{figure}[t]
\centering
\includegraphics[width=\linewidth]{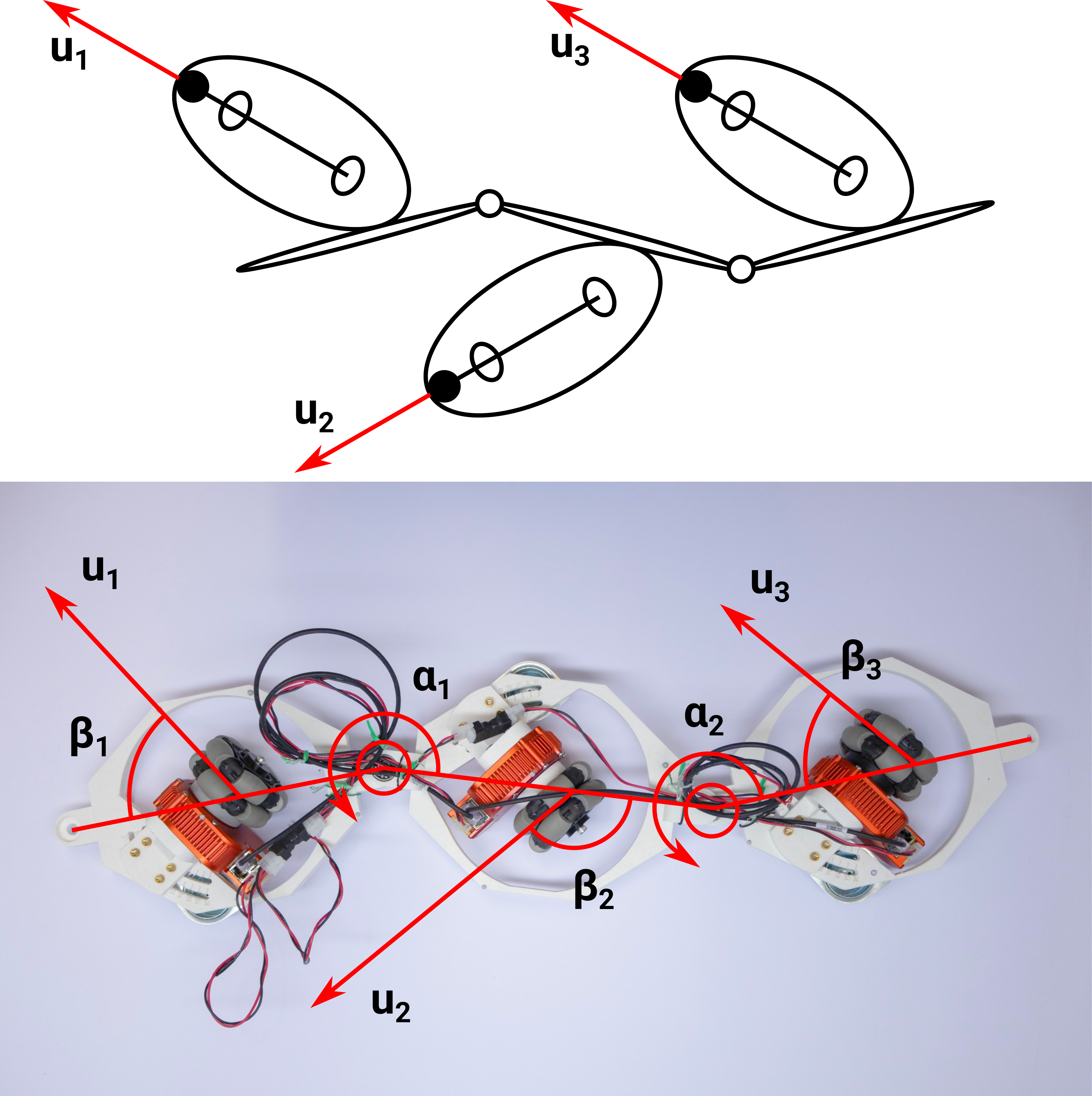}
\caption{Schematic comparison of the LandSalp robot (bottom) and its biological inspiration, the sea salp (top), both with a linear chain configuration. LandSalp consists of three links connected by passive joints, each equipped with a motor-omniwheel assembly that generates thrust and emulates anisotropic viscous drag. The orientation of each assembly relative to its link axis is denoted $\thrustang_i$, with values of $57^\circ$, $130^\circ$, and $57^\circ$ from left to right, and $\ctrl_i$ denotes the corresponding actuator thrust. Joint angles between adjacent links are denoted $\alpha_i$.}
\label{fig:design}
\vspace{-0.5cm}
\end{figure}


Our experimental platform is the LandSalp robot, a planar chain of links equipped with force-controlled omniwheels, shown in Fig.~\ref{fig:design}. This drive system allows us to physically simulate the effects of fluid drag and thrust on a chain immersed in fluid without incurring the logistical complications of operating a system in water. The dynamics of the LandSalp introduce sufficient physical noise to constitute a challenging estimation problem, and the system serves as a proving ground for testing controllers and observers before deployment on a waterborne platform such as FloatSalp, a salp-inspired three-link marine surface robot~\cite{Hecht-2024-development, Choi-2025-gait}.

In previous work, we used external motion capture to perform system identification on the LandSalp and achieve closed-loop control~\cite{Yang2025-ja}. In this paper, we focus on state estimation and closing the feedback loop without reliance on external sensing. The primary contribution is the development and offline demonstration of a sequential state estimator for LandSalp's pose and shape that accounts for non-Gaussian, non-zero-mean process and measurement noise, achieving stable tracking using only onboard gyroscope measurements. Notably, we demonstrate that by exploiting the kinematic constraints of the multi-link chain, the joint angles can be reliably estimated when the system is sufficiently excited by control inputs. This work represents a key milestone toward bridging the gap between controllability and observability in underactuated, minimally sensed multi-link systems of this type.



The remainder of this paper is organized as follows: Section~\ref{sec:platform} introduces the LandSalp robotic platform, Section~\ref{sec:model} develops the corresponding dynamics and measurement models, and Section~\ref{sec:observability} details the design of the sequential state estimator. Section~\ref{sec:results} presents and discusses the offline experimental results, and Section~\ref{sec:conclusion} concludes the paper and outlines directions for future work.

\section{ROBOTIC PLATFORM} \label{sec:platform}
\begin{todolist}
    \item \st{It would be helpful to describe in more detail how the physical equivalence between underwater salp swimming and the terrestrial locomotion of the LandSalp robot was established. This is an especially interesting aspect of the work.}
\end{todolist}
The LandSalp robot, originally designed in~\cite{Hecht2024-design,Yang2025-ja}, is a planar, three-link chain robot that mimics the locomotion of salps on land, as illustrated in Fig.~\ref{fig:design}. The design is motivated by the observation that salps swim in relatively low-Reynolds-number regimes~\cite{Sutherland2017-hydrodynamic}, where the dynamics are dominated by viscous drag. In this regime, the net force and moment acting on each salp unit can be approximated as the sum of thrust and a viscous drag force proportional to body velocity. Leveraging this observation, we established a physical analogy between underwater salp swimming and the terrestrial locomotion of LandSalp: thrust and drag are realized through force control of the series-elastic actuators, while anisotropic viscous drag in water is emulated by the anisotropic friction of the omniwheels. In this way, the robot preserves the key dynamic characteristics of salp locomotion in a planar terrestrial setting. Biological salps are influenced by additional hydrodynamic effects, including added mass, unsteady vortex shedding, and quadratic drag. On the LandSalp platform, these are not modeled explicitly; rather, they correspond to non-ideal terrestrial effects beyond the viscous-friction approximation, such as Coulomb friction, rolling resistance, and cable or harness disturbances.

Each link of the robot is equipped with a 
HEBI series-elastic actuator driving an omniwheel, complemented by a passive ball caster to keep the link level with the ground. Each actuator also carries an onboard IMU, which provides proprioceptive sensing for the robot. Although the IMUs provide both accelerometer and gyroscope measurements, we consider only the gyroscope data in this work. Each link therefore represents a salp unit, with the motor--omniwheel assembly emulating both the propulsion generated by water jets and the viscous drag acting on the body. Because the inter-link joint angles are constrained, translation and rotation of the multi-link system requires cyclic, reciprocating actuation of the omniwheels---referred to as a ``gait''---which is consistent with the pulsatile intake and expulsion of water observed in salps.

The orientation of each omniwheel assembly can be adjusted by mounting it at different positions along a rotation slot, corresponding to different fixed orientations of salp jets relative to the body. In this work, the three omniwheels were mounted at angles of $57^\circ$, $130^\circ$, and $57^\circ$ relative to the link axes, denoted by $\thrustang_i$ and overlaid in Fig. \ref{fig:design}, producing an architecture analogous to the ``linear'' salp colonies in nature.

The LandSalp platform allows us to validate our modeling, estimation, and identification algorithms under controlled conditions with minimal confounding factors. Successful LandSalp experiment results indicate that the proposed algorithms can tolerate meaningful levels of unmodeled dynamics and external disturbances \cite{Yang2025-ja}, providing evidence that the approach can generalize to more field-capable underwater and marine robotic platforms operating in regimes where viscous effects remain dominant.

\section{MODELING AND CONTROL} \label{sec:model}
\begin{todolist}
    \item \st{a description of the system model should be made more clear (addressed in more details) to make the content accessible to a wider range of readers}
    \item \st{The authors should provide the detailed derivation process of Equation (7).}
\end{todolist}


We model the LandSalp robot using a reduced-order geometric mechanics model under a viscous drag-dominated assumption. Unlike a Purcell swimmer, this multi-jet locomotion system has actuation that is not limited to the joint axes. As a result, jet forces project into both the shape and body-motion directions, yielding coupled passive pose and shape dynamics.

\subsection{Dynamics and Measurement Models}
The dynamics of the LandSalp robot are modeled under a quasi-static equilibrium between thrust and drag forces, based on a drag-dominated assumption. This formulation is motivated by both the robot’s design—built to emulate salp locomotion—and by the relatively low-Reynolds-number swimming regime, where viscous drag dominates inertial effects.

We define the system configuration $\bundle$ as the concatenation of shape $\base$ and pose $\fiber$,
\begin{gather*}
\base = \begin{bmatrix}
\jointangle_{1} \\ \jointangle_{2}
\end{bmatrix}, \quad
\bundle =
\begin{bmatrix}
\fiber \\ \base
\end{bmatrix}, \quad
\bundledot =
\begin{bmatrix}
\fibercirc \\ \basedot
\end{bmatrix}, 
\end{gather*}
where $\fiber \in SE(2)$ denotes the planar position and orientation of the robot base, defined at the geometric center and mean orientation of the three wheels; $\base$ contains the shape variables, consisting of the two joint angles $\jointangle_1,\jointangle_2 \in \left[0, 2\pi \right)$; \mbox{$\fibercirc \in \mathfrak{se}(2)$} is the body velocity of the base;\footnote{The ``open circle'' notation used here is similar to the ``dot'' notation, but denotes velocities expressed with local forward and lateral components rather than coordinate-aligned components.%
} and $\basedot$ are the joint velocities.

The body velocity of the $\fstidx$-th salp unit in the chain, $\fibercirc_{\fstidx}$, follows from differential kinematics:
\begin{equation*}
\fibercirc_{\fstidx} =
\inv{\Adj_{\altfiber_{\fstidx}}}\fibercirc +
\sum_{\sndidx}^{\fstidx-1}
\inv{\Adj_{\altfiber_{\frac{\fstidx}{\sndidx}}}}
\begin{bmatrix}0 \\ 0 \\ 1 \end{bmatrix} \dot{\jointangle}_{\sndidx}
= \jac_{\fstidx}^{\text{unit}}\bundledot,
\end{equation*}
where $\altfiber_{\fstidx}(\base)$ is the transformation from the system body frame to the $\fstidx$-th unit, whose body frame is defined at the omniwheel center with the $x$-axis aligned to the controlled spin direction; $\altfiber_{\frac{\fstidx}{\sndidx}}(\base)$ is the transformation from joint $\sndidx$ to unit $\fstidx$; $\Adj_{\bullet}$ denotes the adjoint operator on Lie algebra elements;\footnote{The adjoint operator $\Adj_{\bullet}$ combines the cross-product operation---which converts linear and angular velocities or forces/momenta between different frames---with the rotation operation that expresses velocities or forces/momenta in body-aligned coordinates.} and $\jac_{\fstidx}^{\text{unit}}$ is the Jacobian mapping configuration velocities to the $\fstidx$-th unit’s body velocity. Similarly, we define the Jacobian $\jac_{\sndidx}^{\text{joint}}$ mapping configuration velocities to the \mbox{$\sndidx$-th} joint velocity, and the Jacobian $\jac_{\fstidx}^{\text{imu}}$ mapping them to the $\fstidx$-th IMU’s body velocity, with the IMU frame located at the actuator mounting point on the side opposite the output shaft.

The generalized force on the configuration is
\beq
\label{eq:bundle_force}
\force_{\bundle} = \sum_{\fstidx}^{\fstqty} \jac^{*, \text{unit}}_{\fstidx}
\bigl(\ctrlforce_{\fstidx} + \dragforce_{\fstidx}\bigr) + 
\sum_{\sndidx}^{\fstqty-1} \jac^{*, \text{joint}}_{\sndidx}\dragtorque_{\sndidx},
\eeq
where $\jac^{*}_{\bullet}$ denotes the corresponding dual Jacobian, $\dragforce_{\fstidx}$ and $\dragtorque_{\sndidx}$ are the viscous drag force and torque acting on the units and joints, respectively, and $\ctrlforce_{\fstidx}$ is the thrust of the $\fstidx$-th actuator. Drag is modeled as
\beq
\label{eq:drag_force}
\dragforce_{\fstidx} = -\localmetrics^{\text{unit}}_{\fstidx}\fibercirc_{\fstidx}, \quad
\dragtorque_{\sndidx} = -\localmetrics^{\text{joint}}_{\sndidx}\basedot_{\sndidx},
\eeq
where $\localmetrics^{\text{unit}}_{\fstidx}$ and $\localmetrics^{\text{joint}}_{\sndidx}$ are constant drag matrices defined as
\beq \label{eq:drag_param}
\localmetrics^{\text{unit}}_{\fstidx} =
\begin{bmatrix}
\localmetrics_{\fstidx, xx} & \localmetrics_{\fstidx, xy} & \localmetrics_{\fstidx, x\theta} \\
\localmetrics_{\fstidx, xy} & \localmetrics_{\fstidx, yy} & \localmetrics_{\fstidx, y\theta} \\
\localmetrics_{\fstidx, x\theta} & \localmetrics_{\fstidx, y\theta} & \localmetrics_{\fstidx, \theta\theta}
\end{bmatrix}^{\text{unit}}, \quad
\localmetrics^{\text{joint}}_{\sndidx} = \localmetrics^{\text{joint}}_{\sndidx,\theta\theta}.
\eeq
The thrust generated by the actuators is modeled as
\beq
\label{eq:thrust_force}
\ctrlforce_{\fstidx} =
\begin{bmatrix}
\ctrlforcescalar_{\fstidx} \\ 0 \\ 0
\end{bmatrix} =
\begin{bmatrix}
\localmetrics^{\text{unit}}_{\fstidx, xx}\ctrl_{\fstidx} \\ 0 \\ 0
\end{bmatrix},
\eeq
where $\ctrl_{\fstidx} \in \mathbb{R}$ denotes the commanded velocity. Substituting the thrust and drag force definitions from~\eqref{eq:drag_force} and~\eqref{eq:thrust_force} into the total generalized force expression in~\eqref{eq:bundle_force}, and then collecting the terms affine in the configuration velocity and control separately, yields
\begin{gather*}
\force_{\bundle} = -\Localmetrics\bundledot + \Localmetrics_{\ctrl}\ctrl,
\end{gather*}
with
\begin{align*}
\Localmetrics &= \sum^{\fstqty}_{\fstidx}\jac^{*, \text{unit}}_{\fstidx}\localmetrics^{\text{unit}}_{\fstidx}\jac^{, \text{unit}}_{\fstidx}
+ \sum^{\fstqty-1}_{\sndidx}\jac^{*, \text{joint}}_{\sndidx}\localmetrics^{\text{joint}}_{\sndidx}\jac^{, \text{joint}}_{\sndidx}, \\
\Localmetrics_{\ctrl} &= \sum^{\fstqty}_{\fstidx}\jac^{*, \text{unit}}_{\fstidx, x}\localmetrics^{\text{unit}}_{\fstidx, xx}\transpose{e}_{\fstidx},
\end{align*}
where $e_{\fstidx}$ is the basis vector selecting the $\fstidx$-th control input from $\ctrl$.

For systems in which inertial forces are much smaller than drag forces, the dynamics can be modeled as a quasi-static equilibrium between thrust and drag $\left(\text{i.e.,} \ \force_{\bundle} = \boldsymbol{0}\right)$, with inertial effects assumed to be immediately damped out~\cite{Yang2025-ja, Hatton2013-geometrica, Purcell:1977}:
\begin{align} 
\bundledot &= \underbrace{\inv{\Localmetrics}\Localmetrics_{\ctrl}}_{\cvf(\base)}\ctrl, \label{eq:q_dot}
\end{align}
yielding a first-order dynamics model as a linear map $\cvf(\base)$ from control inputs to configuration velocities.

From the differential kinematics model, one can construct a measurement model for a gyroscope attached to an arbitrary link. Given the Jacobian $\jac_{\fstidx,\text{imu}}$, which maps the system configuration velocity to the velocity of the IMU mounted on the $\fstidx$-th link, the expected gyroscope measurement is modeled as
\beq
\label{eq:gyro}
\omega_i = (\jac_{\fstidx,\text{imu}} \bundledot)_{\theta},
\eeq
where $(\cdot)_{\theta}$ denotes extraction of the rotational component of the body velocity.

\subsection{Gait Design}
Since the linear map $\cvf(\base)$ in \eqref{eq:q_dot} is rank-deficient, not all passive configuration velocities can be controlled instantaneously by a linear combination of thrust inputs. To address this limitation, we adapt the methods proposed in~\cite{Yang2025-ja}, which employ first-order Lie brackets of $\cvf(\base)$ together with averaging theory to design periodic gaits. These gaits satisfy small-time local controllability and realize desired configuration velocities over a gait cycle around a nominal shape.
\begin{figure}[t]
    \vspace{0.1cm}
    \includegraphics[width=\linewidth]{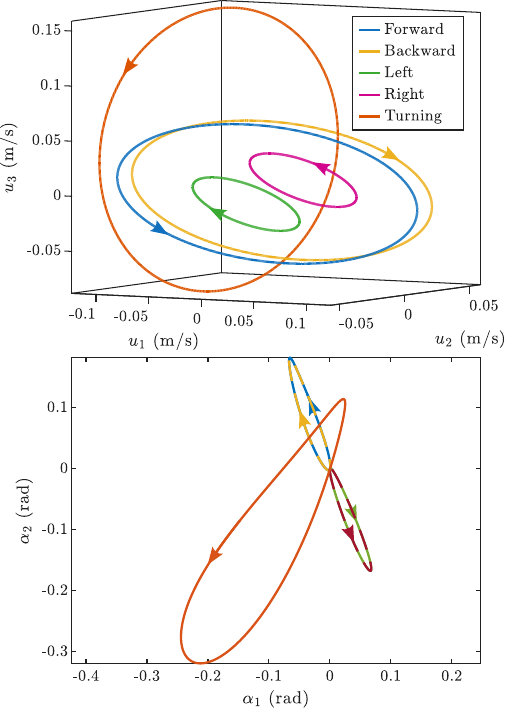}
    \caption{Control (top) and shape trajectories (bottom) for the forward, backward, left, right, and turning gaits.}
    \label{fig:gait_compare}
    \vspace{-0.2cm}
\end{figure}

Each gait is parameterized by a first-order Fourier series, comprising a constant offset and cosine and sine terms. This produces closed cycles in the space spanned by the three control variables, as shown in the top panel of Fig.~\ref{fig:gait_compare}, where gaits corresponding to movements in opposite directions appear symmetrically with respect to the zero-thrust point. Among these motions, the turning gait requires the largest thrust actuation, while the lateral left and right translation gaits require the smallest.

\begin{figure*}[t]
    \vspace{0.1cm}
    \centering
    \includegraphics[width=\linewidth]{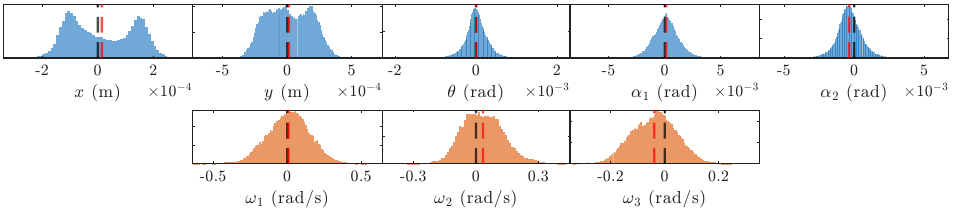}
    \caption{Residual distributions of the process model (top row, blue) and measurement model (bottom row, orange), obtained by comparing the models from Section~\ref{sec:model} against motion capture and gyroscope measurements, respectively. Dashed black and red lines indicate zero and the sample mean, respectively. Residuals were computed from the forward-gait training set at 200 Hz over 45 seconds ($N=9000$).}
    \label{fig:model_resid}
    \vspace{-0.2cm}
\end{figure*}
\section{STATE ESTIMATION} \label{sec:observability}

\begin{todolist}
    \item \st{The proposed estimation scheme requires the selection of several parameters to be effectively implemented. The authors should provide a systematic method for designing or selecting all the parameters required to implement the proposed method. Besides, it would be interesting to evaluate the influence of different parameters on the estimation performance.}
    \item \st{The so-called "Enhanced GP" variant referred to in this paper appears to differ from the GP-UKF framework presented in Reference [16]. The authors should specifically elaborate on the contributions made in this section.}
    \item \st{some comparative study with alternative estimation methods would be well welcomed}
    \item \st{In Section IV, it should be noted that the observability analysis does not refer only to the initial state of the system, but rather to the ability to reconstruct the entire state over time based on the input–output behavior.}
\end{todolist}

Observability describes whether a system's initial state can be uniquely determined from its input–output behavior. Weakly observable or unobservable states typically manifest in a state estimator as unbounded covariance growth and divergence of the state estimate from the true state. Knowledge of a system's observability conditions is therefore a crucial aspect of system design, as it ensures the feasibility of state estimation for safe and efficient feedback control. While a formal nonlinear observability analysis is not presented in this work, prior studies on similar dead-reckoning multi-link systems using IMUs~\cite{Furukawa2021-vb, Andrews2025-it} suggest that this system is observable with active control, assuming an initial pose state is provided. As with any physical dead-reckoning system, drift and slow covariance growth in the position and heading estimates are inevitable, arising from measurement and process model errors accumulated through integration.

While an onboard sequential state estimator relies solely on gyroscope measurements for state updates, motion capture data can be extensively leveraged offline for system identification and design of the state estimator. An initial analysis of model performance, comparing process and measurement model predictions against motion capture and gyroscope data with the experimental setup detailed in Section \ref{sec:results}, revealed that the resulting error residuals were distributed with non-zero mean and non-Gaussian statistics (see Fig.~\ref{fig:model_resid}). While the exact source of these biases is unclear, they are likely attributable to unmodeled dynamics and the inherent non-Gaussian uncertainties arising from representing pose as a Cartesian position and heading~\cite{Long2012-hf}. These residual distributions violate the zero-mean Gaussian assumptions of traditional Kalman filters, motivating the use of data-driven residual learning to correct these discrepancies.

In this work, we employ an Unscented Kalman Filter (UKF) augmented with Gaussian processes (GPs) as the sequential filtering algorithm. GPs are non-parametric, data-driven models with a probabilistic formulation that integrates readily with probabilistic state estimators. The \mbox{GP-UKF} framework was originally introduced in~\cite{Ko2007-db}; here, we employ the ``Enhanced-GP'' variant, in which GPs are used to model the process and measurement residuals rather than the full process and measurement models. This residual-learning formulation was shown in~\cite{Ko2007-db} to outperform both the standard UKF and the conventional \mbox{GP-UKF}, even when trained on sparse data.

Let \(\pfunc\) and \(\mfunc\) denote the true, but unknown, process and measurement functions, respectively, and let \(\hat{\pfunc}\) and \(\hat{\mfunc}\) denote their corresponding approximated representations. For this work, \(\hat{\pfunc}\) and \(\hat{\mfunc}\) are the kinematic and gyroscope models presented in Section \ref{sec:model}. The residual training datasets are defined as  
\begin{align*}
    {D}_{\pfunc} &= \Big\{ (X,U), \; X_+ - \hat{\pfunc}(X,U) \Big\} \\
    {D}_{\mfunc} &= \Big\{ (X,U), \; Y - \hat{\mfunc}(X,U) \Big\},
\end{align*}  
where \mbox{$X = \mtx{\state_1 & \state_2 & \state_3 & \dots}$} is an array of ground-truth states,  
\mbox{$U = \mtx{\ctrl_1 & \ctrl_2 & \ctrl_3 & \dots}$} is an array of control inputs,  
\mbox{$X_+ = \mtx{\pfunc(\state_1,\ctrl_1) & \pfunc(\state_2,\ctrl_2) & \pfunc(\state_3,\ctrl_3) & \dots}$} is an array of ground-truth successor states, and \mbox{$Y = \mtx{\meas_1 & \meas_2 & \meas_3 & \dots}$} is an array of measurements. The general dataset notation is read as $D = \{\text{inputs}, \text{outputs}\}$, with $(X,U)$ denoting concatenated input features. In this work, ground-truth states are measured via motion capture, with time-synchronized gyroscope measurements collected concurrently from each IMU on LandSalp.

A separate GP is trained for each output dimension, yielding five process GPs and three measurement GPs in this application. We use the notation  
$\gpmu{}\big((\state_k,\ctrl_k), D\big)$ to denote the GP predicted mean at $(\state_k,\ctrl_k)$ given training dataset $D$, and $\gpsig{}\big((\state_k,\ctrl_k), D\big)$ for the corresponding predicted variance. Because each GP is trained independently for a single output, the full covariance matrix returned by $\gpsig{}$ is diagonal, with individual variances along the diagonal. The Enhanced-GP-UKF state and measurement prediction steps take the form
\begin{align*}
    \state_{k+1} &= \hat{\pfunc}\!\left(\state_{k}, \ctrl_{k}\right) + \gpmu{}\!\left(\left(\state_{k}, \ctrl_{k}\right), {D}_{\pfunc}\right), \\
    \meas_{k} &= \hat{\mfunc}\!\left(\state_{k}, \ctrl_{k}\right) + \gpmu{}\!\left(\left(\state_{k}, \ctrl_{k}\right), {D}_{\mfunc}\right),
\end{align*}
where the nominal models $\hat{\pfunc}$ and $\hat{\mfunc}$ are augmented by GP mean residual functions. The process and measurement noise covariances are approximated from the GP predicted covariances: 
\begin{align*}
    \qmat_{k+1} &= \gpsig{}\!\left(\left(\state_{k}, \ctrl_{k}\right), {D}_{\pfunc}\right), \\
    \rmat_k &= \gpsig{}\!\left(\left(\state_{k}, \ctrl_{k}\right), {D}_{\mfunc}\right).
\end{align*}  

One limitation of GPs is that their computational complexity scales with the amount of training data and input dimensionality. However, by exploiting the structure of the kinematic geometric model, the process and measurement model GPs can be reformulated with a reduced set of input features depending only on the joint angles, $\base$, and actuator inputs, $\ctrl$. Joint angle constraints and predetermined gaits further reduce the size of the training space. Compared to a traditional UKF, the GP-UKF is more computationally demanding and can present timing challenges for real-time control of fast systems. However, LandSalp and future platforms are expected to exhibit relatively slow dynamics, enabling the estimator to be downsampled (1--10 Hz) from IMU sampling rates (100+ Hz) without significant performance loss. Although more real-time-feasible alternatives exist for residual learning, such as neural networks, the methods presented in this work prioritize successful experimental validation in an offline processing framework.

\section{EXPERIMENTAL RESULTS} \label{sec:results}
\begin{todolist}
    \item Given that the validation through the experimental platform constitutes the main body of this paper, the authors should incorporate additional comparative experiments to elucidate the originality of the state estimation method presented in this paper.
    \item The performance evaluation remains largely qualitative. Providing quantitative performance metrics would strengthen the experimental validation
    \item add control signal alongside estimation results to tie to \cite{Andrews2025-it}
    \item if its not too much work, update Fig. \ref{fig:model_eval} xlabel to 'Time (s)'
\end{todolist}
\begin{figure}[t]
    \centering
    \includegraphics[width=\linewidth]{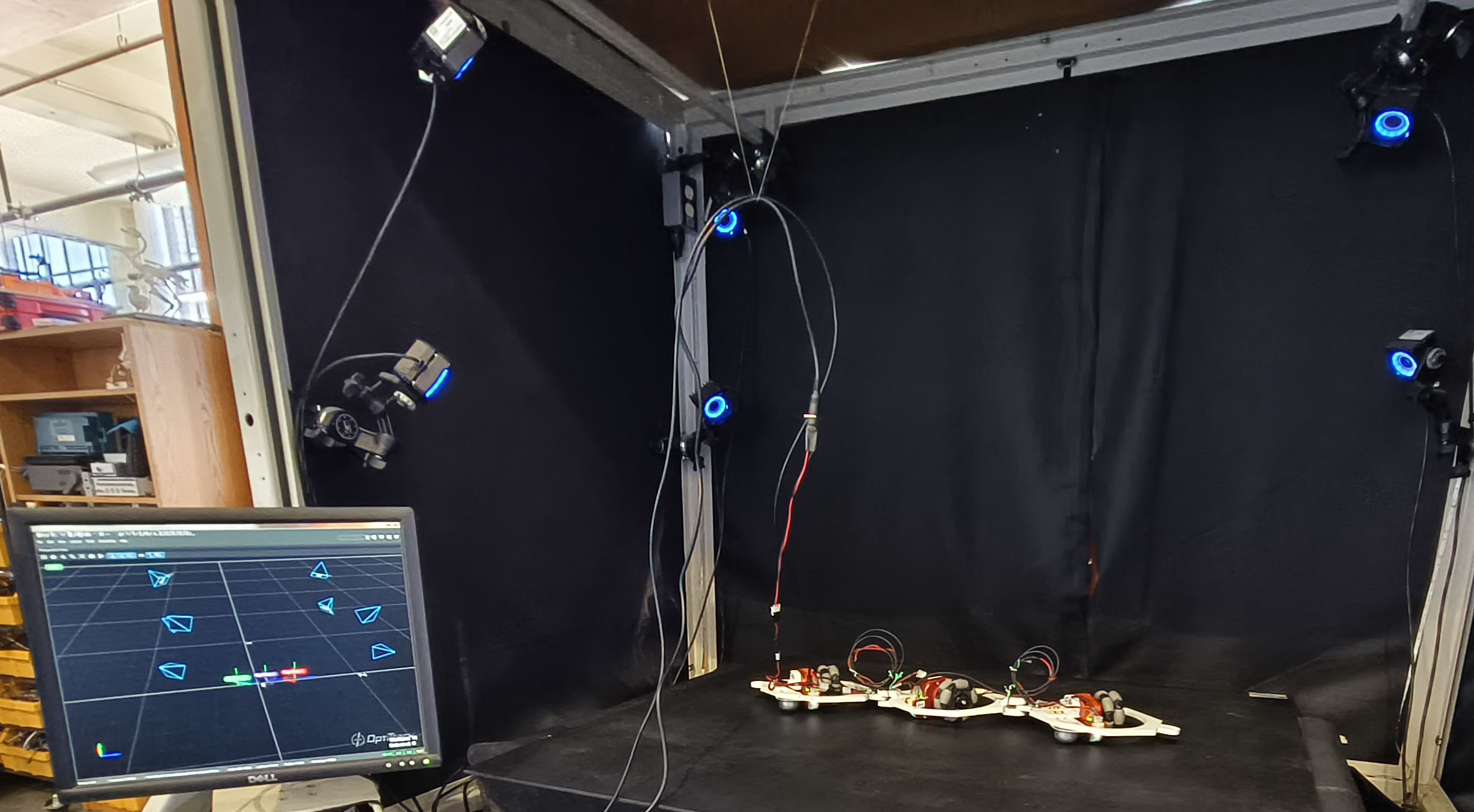}
    \caption{Overview of the LandSalp experimental setup. The robot operates on a platform rigged with an OptiTrack motion capture system. Motion capture data is processed by a desktop computer and transmitted via Robot Operating System (ROS) to a laptop running the gait control algorithm. The robot is tethered by two separate cables for power supply and communication.}
    \label{fig:system_setup}
    \vspace{-0.2cm}
\end{figure}

The LandSalp experimental setup is shown in Fig.~\ref{fig:system_setup}. The experiments were conducted on a flat platform, with robot state logging and feedback control provided by an OptiTrack motion capture system. The motion capture system tracked reflective markers mounted on the robot and streamed pose data to an external laptop at 240 Hz. Control commands were transmitted to each series-elastic actuator at 200 Hz, and the actuators returned actuation states and IMU measurements at the same rate. The robot was powered through lightweight, tension-free tethered cables to avoid introducing external disturbances. Communication and control were managed through Ethernet, with the laptop running ROS Noetic for data handling and controller execution. 


\subsection{Model Validation}
\begin{figure*}[t]
    \vspace{0.1cm}
    \centering
    \includegraphics[width=\linewidth]{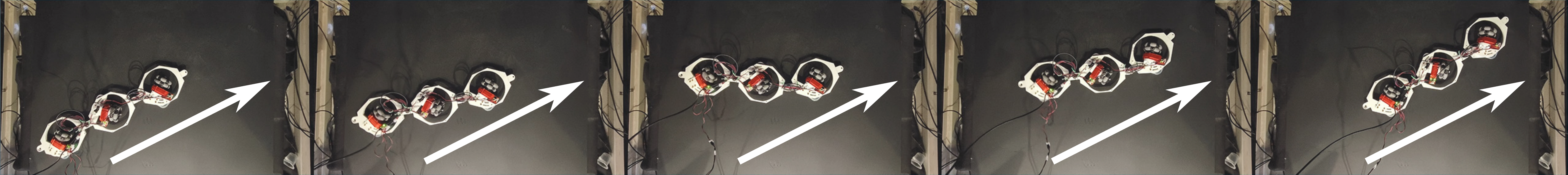}
    \caption{Snapshots of the LandSalp forward-gait experiment showing the robot configuration and position evolving at approximately quarter intervals of the gait cycle, from left to right. White arrows indicate the desired direction of motion.}
    \label{fig:experiment_diagram}
    \end{figure*}
    \begin{figure*}[t]
    \centering
    \includegraphics[width=\linewidth]{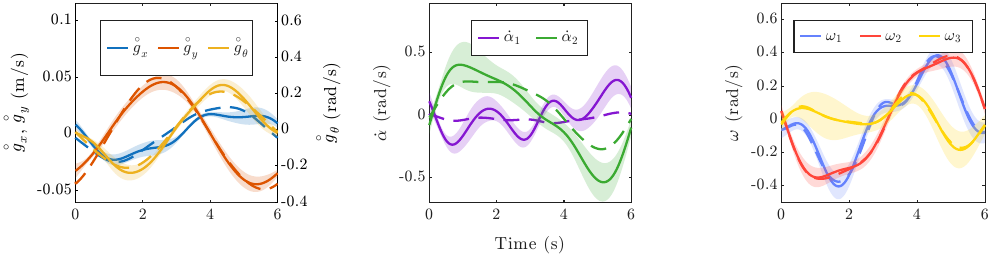}
    \caption{Model validation against measured data, averaged over eight forward gait cycles. (Left, Middle) Estimated pose and shape velocities from motion capture (solid) versus model predictions using motion capture shape estimates and control inputs (dashed). (Right) Gyroscope readings (solid) versus measurement model predictions based on motion capture states (dashed). Shaded regions indicate one standard deviation across gait cycles.}
    \label{fig:model_eval}
    \vspace{-0.2cm}
\end{figure*}
The models from Section~\ref{sec:model} were validated on the \mbox{LandSalp} platform by executing the gaits derived in~\cite{Yang2025-ja}. We selected the forward motion gait as a representative example because it balances position displacement, shape motion, and thrust magnitude, making it well-suited for evaluation and demonstrating the performance of the proposed methods in nontrivial scenarios. The forward gait produces forward translation at a rate of $3$ cm per cycle, with a period of $6$ seconds, while maintaining a nominal configuration of zero joint angles. Snapshots of the forward gait executed on LandSalp are shown in Fig.~\ref{fig:experiment_diagram}.

The collected data were post-processed before analysis. The motion capture system provided the configuration states of the robot, from which velocities and accelerations were approximated using groupwise finite differences~\cite{Yang2025-ja}. To mitigate the high-frequency noise introduced by numerical differentiation, a low-pass filter was applied to the measured trajectories. Similarly, the gyroscope signals were low-pass filtered to suppress sensor noise. The choice of filter cutoff frequencies was informed by the executed gait frequency and the measured noise characteristics, exploiting the fact that the system operates in an overdamped regime and starts and ends at rest. 

The drag parameters \eqref{eq:drag_param} in the dynamics model \eqref{eq:q_dot} were identified from post-processed experimental training data via linear regression with physical consistency constraints, following the methods in~\cite{Yang2025-ja}. The left and middle panels of Fig.~\ref{fig:model_eval} validate the dynamics model by comparing its prediction of passive configuration velocity—given shape measurements from motion capture and the applied control input—with the velocity measured by motion capture during the forward motion experiment. The data were averaged over all cycles, and shaded regions indicate the standard deviations across cycles. The model’s predictions capture the dominant trends in both position and shape velocities. Predictions of position velocity are more accurate than those of shape velocity; the latter exhibit second-order effects resembling spring-like behavior from the harness on the robot, which are beyond the scope of the first-order model. Nevertheless, the model accurately reproduces the primary trends in joint velocities.



The measurement model was also validated experimentally by comparing measured gyroscope data with the measurement model predictions evaluated from motion capture-derived states. Because the motion capture measurements serve as a near-ground-truth approximation of the system state, this comparison provides an independent check on model accuracy. The right panel of Fig.~\ref{fig:model_eval} shows a representative comparison for the second link. The gyroscope data closely matches the predicted values, with variations primarily attributed to sensor noise and vibrations introduced from the actuators and omniwheel rollers during contact with the platform.

\subsection{State Estimator Parameter Selection}
To evaluate the dependence of GP-UKF performance on the training data distribution, two GP training datasets are considered: one consisting solely of forward-gait experiments, and the other comprising an equal distribution of all five gaits, referred to as ``multi-gait.'' Both datasets contain 2250 data points collected over 450 s.

In both training scenarios, unfiltered motion capture and gyroscope measurements were downsampled to 5 Hz to reduce dataset size while preserving the overall system behavior. A total of eight GPs were trained --- five modeling process residuals (one per state variable) and three modeling measurement residuals (one per gyroscope) --- each with a five-dimensional input space comprising two joint angles and three actuator inputs. All GPs were trained using MATLAB's Gaussian process regression function (\texttt{fitrgp}) with default hyperparameter settings, optimizing over kernel variance and data normalization for 30 iterations. Rather than pursuing exhaustive hyperparameter tuning, this work intentionally adopts a minimal training procedure to demonstrate the robustness of the GP-UKF framework under practical training conditions.

\begin{figure*}[t]
    \vspace{0.1cm}
    \centering
    \includegraphics[width=\linewidth]{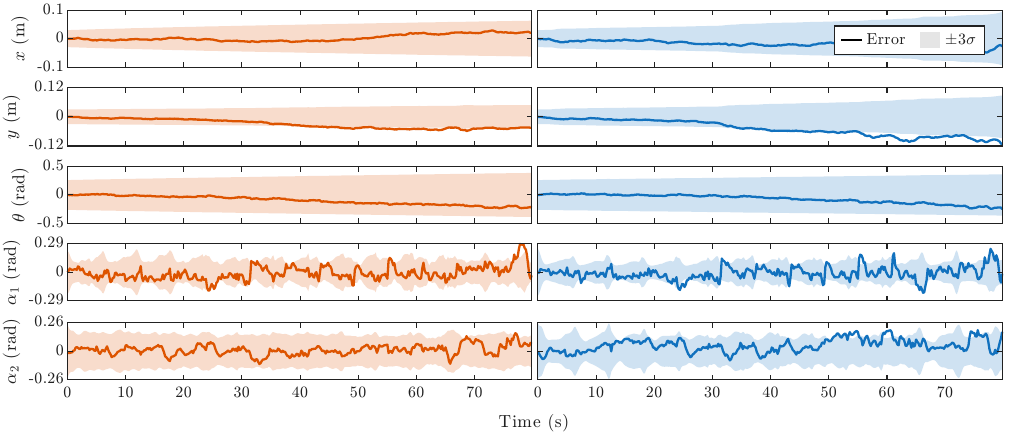}
    \caption{Pose estimation errors relative to motion capture ground-truth with $\pm 3\sigma$ covariance bounds for the forward gait (left column, orange) and multi-gait (right column, blue) GP-UKFs, both evaluated on the same held-out forward-gait test trajectory.}
    \label{fig:kf_pose}
    \vspace{-0.2cm}
\end{figure*}

After training on their respective datasets, the forward and multi-gait GP-UKFs were evaluated offline on the same forward-gait trajectory that was held out from the training datasets. Both GP-UKFs were initialized with identical initial state estimates taken directly from the motion capture system at the time of initialization. To evaluate estimator performance under degraded initialization conditions, initial estimate covariances were set conservatively, with position, orientation, and joint angle standard deviations of 1 cm, $5^\circ$, and $5^\circ$, respectively. Tracking results for both filters are shown in Fig.~\ref{fig:kf_pose}, with the estimate error calculated by comparing the estimates against the motion capture states.


\subsection{Discussion}
Both filters exhibit slow drift in position and orientation estimates, which is expected for dead-reckoning systems where position is integrated rather than directly observed, and can be corrected through occasional localization updates, such as from beacons or GPS. For joint angle states, both filters demonstrate the feasibility of accurate estimation without angle encoders. The comparable performance of both filters highlights that overlapping state and input pairs across gaits can be exploited to reduce per-gait training data requirements while generalizing across multiple gaits. When evaluated on the remaining gaits, the multi-gait filter performed similarly across forward, backward, left, and right gaits, with modest degradation on the turning gait — exhibiting larger covariance bounds, greater drift, and more frequent $3\sigma$ violations, likely due to the larger thrust inputs and joint angle range traversed. We are confident that both the intermittent $3\sigma$ exceedances and pose drift can be mitigated through additional GP training with sufficient coverage of the shape and input space.

A noteworthy feature of the state estimator is the periodic increase-decrease pattern observed in the joint angle estimate covariances. While the observability conditions derived in~\cite{Andrews2025-it} were developed for a higher-order version of this system incorporating accelerometer measurements, we hypothesize that analogous conditions exist here, specifically, that the system is observable when control is active and that observability scales with input magnitude. We therefore attribute the periodic covariance behavior to the sinusoidal gait inputs, whose varying phase and magnitude periodically excite the system and render the joint angles more observable.

To further investigate this hypothesis, we construct an observability heuristic based on the Jacobian of the measurement with respect to the state $\left(\frac{\partial h}{\partial x}\right)$. This Jacobian is significant because it appears as the zeroth-order Lie derivative in the observability codistribution matrix for nonlinear systems~\cite{Hermann1977-rt} and in the observability Gramian of LTV systems
\begin{align*}
    W_O(t_0, t_1) &= \int_{t_0}^{t_1} {\Phi}(\tau, t_0)^\top C(\tau)^\top C(\tau) {\Phi}(\tau, t_0) \, d\tau,
\end{align*}
where $\quad C(t) = \left.\frac{\partial h}{\partial x}\right|_{x(t), u(t)}$ is the Jacobian of the nonlinear measurement function $h(x, u)$ linearized about a nominal trajectory $x(t), u(t)$, and $\Phi(t, t_0)$ is the state transition matrix of the linearized dynamics.

To derive the heuristic for this system, we first rewrite the concatenated measurement function for the three gyroscopes on LandSalp using \eqref{eq:gyro}, with the dependence on $\base$ and $\ctrl$ made explicit by substituting \eqref{eq:q_dot}:
\begin{align*}
    \measurementmodel(\base, \ctrl) &= \mtx{\omega_1(\base, \ctrl)  & \omega_2(\base, \ctrl)  & \omega_3(\base, \ctrl)}.
\end{align*}
The measurement function is control-affine and can be written as
\mbox{$\measurementmodel(\base, \ctrl) = \sum_{i=1}^3 \measurementmodel_{i}(\base) \ctrl_i,$}
where $\measurementmodel_{i}(\base) \in \mathbb{R}^3$ is the contribution of $\ctrl_i$ to $\measurementmodel(\base, \ctrl)$. Focusing on the first joint angle, the Jacobian of the measurement with respect to $\jointangle_1$ is
\begin{gather*}
    \pd{\measurementmodel(\base, \ctrl)}{\jointangle_1} = \sum_{i=1}^3 \pd{\measurementmodel_i(\base)}{\jointangle_1} \ctrl_i,
\end{gather*}
where the control-affine structure allows the Jacobian to be decomposed into per-actuator contributions.

We define the following observability heuristics based on norms and projections of the measurement Jacobian:
\begin{gather*}
    \Lambda = \left\Vert \pd{\measurementmodel(\base, \ctrl)}{\jointangle_1} \right\Vert, \quad 
    \lambda_i = \frac{\left\langle \pd{\measurementmodel(\base, \ctrl)}{\jointangle_1}, \pd{\measurementmodel_{i}(\base)}{\jointangle_1} \ctrl_i \right \rangle}{\left\Vert \pd{\measurementmodel(\base, \ctrl)}{\jointangle_1} \right\Vert},
\end{gather*}
where $\Lambda$ represents the total observability heuristic and $\lambda_i$ is the contribution of actuator $i$, with $\Lambda = \sum_{i=1}^3 \lambda_i$. Note that $\lambda_i$ is the projection of $\pd{\measurementmodel_{i}(\base)}{\jointangle_1} \ctrl_i$ onto $\pd{\measurementmodel(\base, \ctrl)}{\jointangle_1}$. The $\Lambda$ heuristic is noteworthy because it is the square root of the diagonal element of $C(t)^\top C(t)$ corresponding to $\alpha_1$, and generalizes to any state element of interest.

\begin{figure}[t]
    \vspace{0.1cm}
    \centering
    \includegraphics[width=\linewidth]{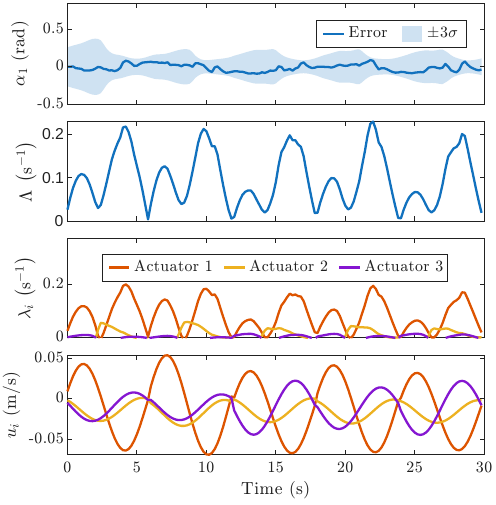}
    \caption{Joint angle estimation results for a multi-gait filter tracking the left gait. From top to bottom: estimation error with respect to motion capture ground truth with $\pm 3\sigma$ covariance bounds; $\Lambda$ observability heuristic; per-actuator $\lambda_i$ heuristics; and commanded velocity.}
    \label{fig:kf_left}
    \vspace{-0.2cm}
\end{figure}
The motion captured measurement of joint angle, the estimate error and covariance, and the observability heuristics are shown in Fig.~\ref{fig:kf_left} for the left gait, which maintains a nearly constant heading while translating laterally approximately 30 cm. The heuristics are evaluated using motion capture states and commanded control inputs. In general, a larger $\Lambda$ corresponds to a more observable state and input configuration, resulting in a smaller, more confident estimate covariance. However, it appears that a minimum magnitude of $\Lambda$ is necessary to overcome sensor and process noise before a noticeable effect on the estimate covariance is observed. A small time delay of less than one second between $\Lambda$ peaks and covariance minima is also visible, which we attribute to delays between commanded and executed control inputs and the transient dynamics of the covariance.

Inspection of $\lambda_i$ reveals that actuator 1 contributes most to $\Lambda$, which is consistent with the gait structure and being attached to an end link unconstrained by a joint on one side affords it greater influence over the system's configuration, an effect proportional to the mass and length of the link. The periodic behavior of both heuristics reflects the periodicity of the gait. These qualitative results support the hypothesis that observability is related to the control input, consistent with the conditions found in~\cite{Andrews2025-it}, and motivate further investigation into observability-optimal gait design.

\section{CONCLUSION AND FUTURE WORK} \label{sec:conclusion}
\begin{todolist}
    \item the results are entirely offline, and while the authors acknowledge this limitation, a clearer outlook on the expected challenges for real-time implementation would enhance the paper's impact.
    \item a discussion on generalization limitations of the proposed methodology, and on difficulties of its online applications is expected
    \item the paper would benefit from a clearer discussion of the limits of generalisability of the approach and a more critical comparison with existing state estimation methods for minimally sensed multi-link systems.
\end{todolist}
The results presented here mark an important step in demonstrating the joint controllability and observability of this new class of underactuated, minimally sensed multi-link robots. 
Despite these promising results, several computational challenges remain before this offline proof of concept can be realized as a robust, onboard real-time system. In the near term, transitioning the GP training pipeline and filter implementation to Python or C++ would allow faster training on larger datasets, enhancing both filter accuracy and processing speed. Additionally, as multi-link systems scale to tens or hundreds of links, the state space grows significantly in dimension, rendering the GP input space increasingly difficult to cover effectively. Replacing the non-parametric GP model with a parametric alternative, such as a neural network or a Bayesian linear regression model, could alleviate this scaling burden. A more systematic investigation of gait-space similarities could further reduce redundant training data while ensuring adequate coverage of less-represented input regions.

In the long term, we plan to conduct experiments on FloatSalp. This salp-inspired three-link marine surface robot will facilitate the estimation of fluid added-mass effects and provide deeper insights into gait design, control strategies, and state estimation for salp-inspired multi-link marine robots~\cite{Hecht-2024-development, Choi-2025-gait}. More broadly, we envision extending these methods and the inertial parameter estimation results of~\cite{Andrews2025-it} to multi-link soft robotic systems, where modeling, estimation, and control remain active areas of research.

\bibliography{references}

\end{document}